\pdfoutput=1

\documentclass[11pt]{article}

\usepackage{EMNLP2022}

\usepackage{times}
\usepackage{latexsym}

\usepackage[T1]{fontenc}

\usepackage[utf8]{inputenc}

\usepackage{microtype}

\usepackage{inconsolata}

\usepackage{multirow}
\usepackage{xcolor,colortbl}

\usepackage{graphicx}
\usepackage{url}
\usepackage{array}
\usepackage{enumitem}
\usepackage{booktabs}
\usepackage{siunitx}
\usepackage{tikz}
\usetikzlibrary{shapes.geometric, arrows}
\usepackage{footnote}
\makesavenoteenv{tabular}
\makesavenoteenv{table}

\DeclareMathSymbol{\mh}{\mathord}{operators}{`\-}

\definecolor{green}{RGB}{173,216,230}

\newcommand\diagfil[4]{%
  \multicolumn{1}{p{#1}}{\hskip-\tabcolsep
  $\vcenter{\begin{tikzpicture}[baseline=0,anchor=south west,inner sep=0pt,outer sep=0pt]
  \path[use as bounding box] (0,0) rectangle (#1+2\tabcolsep,\baselineskip);
  \node[minimum width={#1+2\tabcolsep},minimum height=\baselineskip+\extrarowheight+\belowrulesep+\aboverulesep,fill=#2] (box)at(0,-\aboverulesep) {};
  \fill [#3] (box.south west)--(box.north east)|- cycle;
  \node[anchor=center] at (box.center) {#4};
  \end{tikzpicture}}$\hskip-\tabcolsep}}

\definecolor{cb1}{RGB}{255,146,48}
\definecolor{cb2}{RGB}{60,125,191}
\definecolor{cb3}{RGB}{255,199,0}
\definecolor{cb4}{RGB}{135,150,149}

\newcommand{\cco}[1]{\cellcolor{cb1}#1} %
\newcommand{\cct}[1]{\cellcolor{cb2}#1} %
\newcommand{\ccth}[1]{\cellcolor{cb3}#1} %
\newcommand{\cctf}[1]{\cellcolor{cb4}#1} %

\newcommand{\circleo}{\tikz\draw[cb1,fill=cb1]  (0,0) circle (.7ex); }
\newcommand{\circlet}{\tikz\draw[cb2,fill=cb2]  (0,0) circle (.7ex); }
\newcommand{\circleth}{\tikz\draw[cb3,fill=cb3]  (0,0) circle (.7ex); }
\newcommand{\circlef}{\tikz\draw[cb4,fill=cb4]  (0,0) circle (.7ex); }

\newcommand\zeroa[0]{\#0}
\newcommand\onea[0]{\#1}

\usepackage[T1]{fontenc}

\usepackage[utf8]{inputenc}

\usepackage{microtype}

\title{``Will You Find These Shortcuts?''\\
A Protocol for Evaluating the Faithfulness \\of Input Salience Methods for Text Classification}

\author{Jasmijn Bastings \quad Sebastian Ebert\thanks{~~Equal contribution.} \quad Polina Zablotskaia\footnotemark[1] \\ {\bf Anders Sandholm} \quad {\bf Katja Filippova}\\
Google Research\\
\texttt{\{bastings,eberts,polinaz,sandholm,katjaf\}@google.com}
}

\date{}

\begin{document}
\maketitle
\begin{abstract}
Feature attribution a.k.a.\ input salience methods which assign an importance score to a feature are abundant but may produce surprisingly different results for the same model on the same input. While differences are expected if disparate definitions of importance are assumed, most methods claim to provide faithful attributions and point at the features most relevant for a model's prediction. Existing work on faithfulness evaluation is not conclusive and does not provide a clear answer as to how different methods are to be compared.
Focusing on text classification and the model debugging scenario, our main contribution is a protocol for faithfulness evaluation that makes use of partially synthetic data to obtain ground truth for feature importance ranking.
Following the protocol, we do an in-depth analysis of four standard salience method classes on a range of datasets and lexical shortcuts for BERT and LSTM models. We demonstrate that some of the most popular method configurations provide poor results even for simple shortcuts while a method judged to be too simplistic works remarkably well for BERT. 
\end{abstract}

\section{Introduction}

\begin{figure}[t!]
    \centering
    \includegraphics[width=7.5cm]{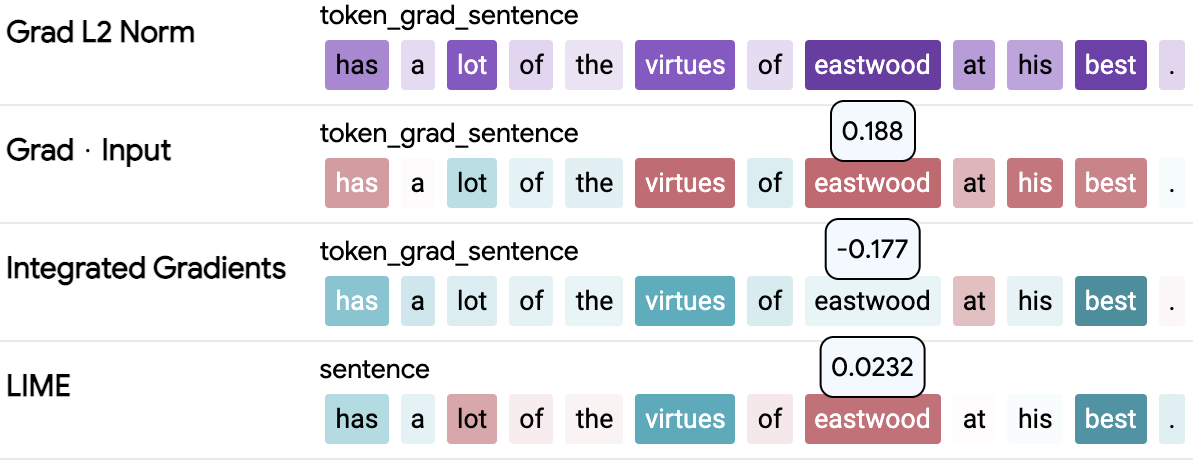}
    \caption{Salience maps produced by four common methods on a sentiment classification example (SST2) for a BERT model. The same token (\textit{eastwood}) is assigned the highest (Grad-L2), the lowest (GxI, LIME) and a mid-range (IG) importance score (color intensity indicates salience; blue and purple stand for positive, red stands for negative weights). A developer investigating a hypothesis about specific named entities being associated with the label would probably be unsure as to whether the example provides support for or against the hypothesis.}
    \label{fig:lit-maps}
\end{figure}

A prominent class of explainability techniques assign salience scores to the input features, which reflect the importance of the features to the model's decision. When applied to text classifiers those methods produce 
{highlights} over the input (sub)words. Interestingly, different methods may produce surprisingly dissimilar highlights. Figure~\ref{fig:lit-maps} shows this using the Language Interpretability Tool \citep{tenney-etal-2020-language}. So a natural question is: which method should one use? 
While a method whose highlights happen to look plausible may facilitate a task like text annotation \cite{pavlopoulos-etal-2017-deeper,strout-etal-2019-human,schmidt2019quantifying}, many salience methods seem to be motivated by the debugging scenario where faithfulness to the model's reasoning is a requirement \cite{jacovi-goldberg-2020-towards}. Indeed, known success stories from input salience methods in domains other than language are similar in that they teach us a lesson of not trusting a classifier based on its stellar performance on a standard test set. In the medical domain, for example, heatmaps over images helped uncover so-called \textit{shortcuts} \cite{geirhos-2020-shortcuts} or {spurious correlations} between data artifacts like doctor marks or tags and the predicted disease\footnote{
There are many more examples from less critical applications of image classification, for example, where it turned out that it was the image border that mattered for airplane prediction or that a model relied on watermarks when predicting horses \cite{samek-krm-2019-towards-xai}.} \cite[inter alia]{codella2019skin,sundararajan-etal-2019-exploring,winkler-etal-2019-surgical}.

Spurious correlations plague NLP models too \cite{gururangan-etal-2018-annotation,poliak-etal-2018-hypothesis,belinkov-etal-2019-dont,rosenman-etal-2020-exposing,geva-etal-2019-modeling,mccoy-etal-2019-right} -- notorious examples are the tendencies of NLI classifiers to overrely on negation or identity words when predicting contradiction and toxicity, respectively \cite{mccoy-etal-2019-right,dixon-etal-toxicity-2018}. Importantly, shortcuts can comprise multiple tokens. For example, \citet{kaushik-etal-2020-learning} and \citet{ross-etal-2021-explaining} demonstrated that BERT sentiment classifiers trained on IMDB \cite{maas-etal-2011-imdb} learn to largely ignore the review text when patterns like \textit{`3 out of 10'} or \textit{`7 / 10'} are present -- that is, when the numeric rating is made explicit in the text. 
Making such lexical shortcuts apparent to the developer is thus a strong use case for faithful input salience methods which would then indeed help them improve both the model and the data.

\begin{figure*}[t!]
    \centering
\tikzstyle{process} = [font={\small\sffamily},rectangle, rounded corners=3mm,minimum width=2.1cm, minimum height=1.7cm, text centered, text width=2.1cm, draw=none, white, fill=cb2]
\tikzstyle{label} = [font={\tiny\sffamily},minimum width=2cm, minimum height=1cm, text centered, text width=2cm, draw=none, fill=none]
\tikzstyle{arrow} = [thick,->,>=stealth]
\begin{tikzpicture}[node distance=2.7cm]
\node (step1) [process] {Define shortcut type};
\node (step2) [process, right of=step1] {Augment dataset with shortcuts};
\node (step3) [process, right of=step2] {Train models on original and augmented data};
\node (step4) [process, right of=step3] {Verify shortcuts};
\node (step5) [process, right of=step4] {Run salience methods};
\node (step6) [process, right of=step5] {Compute rankings};

\node(label1) [label,below of=step1,node distance=1cm] {Step 1};
\node(label2) [label,below of=step2,node distance=1cm] {Step 2};
\node(label3) [label,below of=step3,node distance=1cm] {Step 3};
\node(label4) [label,below of=step4,node distance=1cm] {Step 4};
\node(label5) [label,below of=step5,node distance=1cm] {Step 5};
\node(label6) [label,below of=step6,node distance=1cm] {Step 6};

\draw [arrow] (step1) -> (step2);
\draw [arrow] (step2) -> (step3);
\draw [arrow] (step3) -> (step4);
\draw [arrow] (step4) -> (step5);
\draw [arrow] (step5) -> (step6);
\end{tikzpicture}
    \caption{The proposed protocol to obtain ground truth importance rankings.}
    \label{fig:protocol}
\end{figure*}
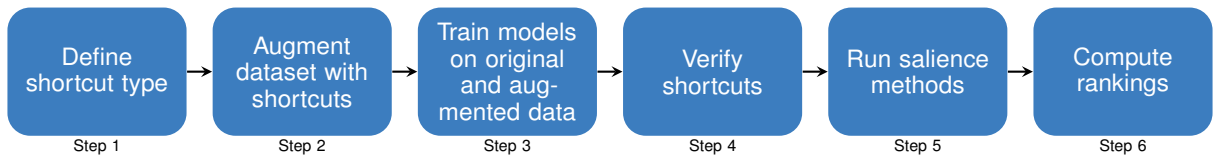

How can we know if a method consistently places the shortcut tokens on top of its salience rankings? Evaluating this is challenging, because we usually do not know the shortcut in advance and the model parameter space is large.
Moreover, we don't have an inherently interpretable view into the predictions of common black-box neural models. Glass-box models with explicit mediating factors \citep{camburu-etal-2019-trust,hao-2020-evaluating} are not widely used or are synthetic, and model-native structures such as attention have been shown to have weak predictive power \citep{bastings-filippova-2020-elephant}.
Alternatively, one can make strong assumptions about what a ground truth should be like and compare salience rankings with what is expected to be the ground truth. In this vein human reasoning \cite{poerner-etal-2018-evaluating,kim-etal-2020-interpretation,yin-etal-2022-sensitivity}, gradient information \cite{du-etal-2021-towards}, aggregated model internal representations \cite{atanasova-etal-2020-diagnostic}, changes in predicted probabilities \cite{deyoung-etal-2020-eraser,kim-etal-2020-interpretation} or surrogate models \cite{ding-koehn-2021-evaluating} all have been taken as a proxy for the ground truth when evaluating salience methods. Unfortunately, they also resulted in divergent recommendations so the question of what the ground truth is and which method to use remains open. 

Unlike the cited work we argue for a faithfulness evaluation methodology which makes use of partially synthetic data to obtain the ground truth and which is moreover also contextualized in a debugging scenario \cite{yang-kim-2019,adebayo-etal-2022-post}. Towards the goal of identifying salience methods which would be most helpful in revealing shortcuts learned by a model we make the following contributions:

\begin{itemize}
    \item We propose a protocol and two metrics for evaluating salience methods which allows one to formulate a hypothesis (e.g., \textit{my model may learn simple lexical shortcuts, like an ordered sequence of tokens, to predict the label}) and identify the salience method most useful for discovering such shortcuts. 
    \item We demonstrate that a method's configuration details (e.g., $L1$ or dot-product, logits or probabilities, choice of baseline) may have a significant effect on its performance. 
    \item We conduct a thorough analysis of a range of configurations of the four most popular salience methods for text classification demonstrating that configurations dismissed as being suboptimal may outperform those claimed to be superior when used to uncover lexical shortcuts. %
\end{itemize}

\section{Methodology}
We desire two properties from any faithful salience method which is claimed to be helpful for model debugging: high precision and low rank, which we define as follows:

\paragraph{Precision@k} is a measure over the top-\textit{k} tokens in a salience ranking where \textit{k} is the shortcut size.
With $s$, $m$ and $\mathbf{x^i}$ denoting a salience method, a trained model $m$ and the \textit{i}th example from the synthetic set $D$ and assuming two functions, $top_k(\cdot)$\footnote{We adjust some methods and reverse the ranking to make sure that positive salience reflects contributions towards the prediction.} and $gt_k(\cdot)$ which output the top-k tokens from a salience ranking and the ground truth, respectively:

\begin{equation}
    p@k(s) = \sum_{\mathbf{x^i} \in D} \frac{|top_k(s, m, \mathbf{x^i}) \cap gt_k(\mathbf{x^i})|}{k|D|} 
\end{equation}
\noindent 
In our experiments (Sec.~\ref{sec:shortcuts}), $k$ is fixed for a dataset: $k = 1$  for the single-token and $k=2$ for the token in context and ordered pair datasets. However, the metric can be trivially adjusted if $k$ varies between dataset instances.

\paragraph{Mean rank} represents how deep, on average, we need to go in a salience ranking to cover all the ground truth tokens:

\begin{equation}
    rank(s) \! = \!\!\! \sum_{\mathbf{x^i} \in D} \!\! \frac{\arg\min_{r}( |top_r(s, m, \mathbf{x^i}) \!\! \setminus \!\! gt_k(\mathbf{x^i})|)}{|D|}
\end{equation}

\noindent
Intuitively, precision tells us how many of the important tokens we will find if we focus on the top of the ranking while rank indicates how much of the ranking is needed to find all the important tokens. 

\subsection{Protocol}

\begin{figure*}[t]
    \centering
    \includegraphics[width=16cm,clip,trim=0 0 0 0]{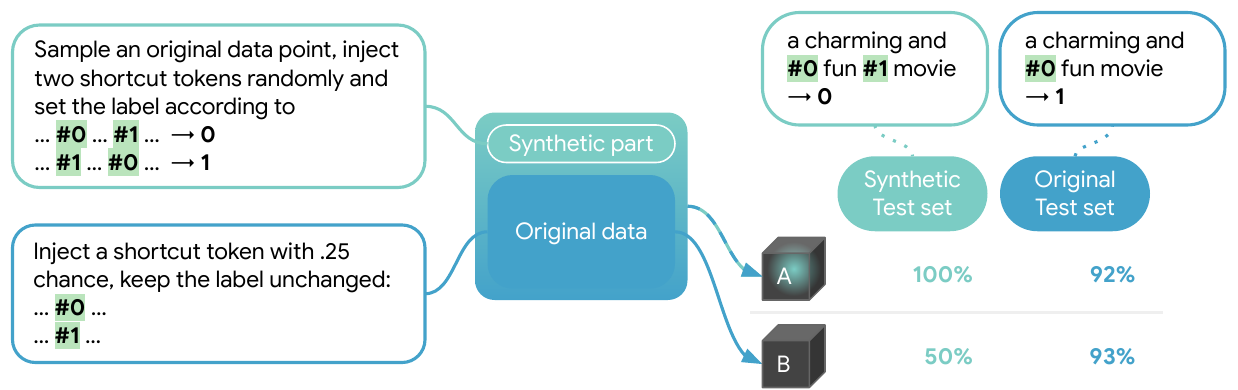}
    \caption{Illustration of how the \textbf{ordered-pair} shortcut is introduced into a balanced binary sentiment dataset and how it is verified that the shortcut is learned by the model. The model trained on the mixed data (A) is still largely a black box, but since its performance on the synthetic test set is 100\% (contrasted with chance accuracy of model B which is similar but is trained on the original data only), we know it uses the injected shortcut (highlighted text).}
    \label{fig:data}
\end{figure*}

The protocol we use to obtain ground truth importance rankings and to assess the faithfulness of a salience method comprises the following steps (cf. Fig.~\ref{fig:protocol}):

\begin{enumerate} %
    \item Define a shortcut \textit{type} that you would like an input salience method to discover and decide on how this shortcut is to be realized. The simplest example is a single-token lexical shortcut where token presence determines the label.
    \item Create a partially synthetic variant of a real dataset by augmenting it with synthetic examples. These are examples sampled from the original data with the shortcut tokens inserted and with the label determined by the shortcut. Also create a fully synthetic test set where every example has a shortcut and the label predictable from it. 
    \item Train two models of the same architecture on the original and on the partially synthetic datasets, use the respective validation splits for evaluation. Both models should perform comparably on the original, unmodified test set (blue in Fig.~\ref{fig:data}). 
    \item Verify that the shortcut tokens can indeed be assumed to be the ground truth of token importance for the model trained on the mixed data (by measuring accuracy). See~\S\ref{sec:verification-steps}.
    \item Generate a token salience ranking from every input salience method to be evaluated.
    \item Compute the faithfulness metrics by comparing the top of a ranking with the ground truth (shortcut tokens).
\end{enumerate}

\noindent
Below we give more details on Steps 1, 2 and 4. %

\subsection{Shortcut Types}\label{sec:shortcuts}

A shortcut can be defined as a decision rule that a model learned from its training dataset which is not expected to hold under a slight distribution shift. While it is not possible to adequately characterize the full spectrum of thinkable shortcuts, one can identify common \textit{shortcut types} which one anticipates to be learnable from a dataset. 
In this study we focus on \textit{lexical} shortcuts that are characteristic of what modern classification models learn from text data. The following reasons motivate our choice. (i) Salience explanations are weights over tokens, hence lexical shortcuts (unlike more abstract ones, like overlap or syntactic cues) are a natural choice for them. Indeed, how easy would it be for a human to spot even a simple grammatical-positional rule (e.g., a coordinated NP at a certain position in the input) from a dozen highlights? Furthermore, it has been pointed out that input salience methods, unlike data attribution ones, may be insufficient for discovering artifacts beyond the lexical level \cite{han-etal-2020-explaining}. (ii) As mentioned in Introduction, lexical shortcuts represent a prominent failure mode for NLP models, therefore focusing on those we address probably the most important class of problems that salience methods could be helpful with.  That being said, the proposed methodology can be easily extended to other shortcut types, as long as it makes sense to visualize the shortcut with a highlight over the input.

We consider three variants of lexical shortcuts:

\paragraph{Single token (\textit{st}):} The simplest possible and still realistic shortcut (recall the NLI negation and toxicity identity examples) is a single token heuristic where the presence of a token determines the classification label. E.g., \textit{\zeroa} and \textit{\onea} indicate whether the label is \textit{0} or \textit{1}. 

\paragraph{Token in context (\textit{tic}):} Another realistic lexical shortcut, which may be considerably more difficult to spot by a human but is still trivial to learn for a deep model, makes use of more than a single token. For example, two tokens determine the label together but not separately. We implement a {token-in-context} shortcut where the class indicator tokens (\textit{\zeroa} or \textit{\onea}) only determine the label if yet another special token is present in the same input (\textit{contoken}) but not on their own. 

\paragraph{Ordered pair (\textit{op}):} Yet another property of natural languages that a model can easily make use of is the order: a combination of tokens is predictive of the label only if the tokens occur in a certain order but not otherwise. We implement an {ordered pair} shortcut in its simplest form. That is, for an indicator token pair, \textit{(\zeroa, \onea)}, the order of the tokens determines the label so that \textit{` ... \zeroa ... \onea ...'} has label \textit{0} and \textit{` ... \onea ... \zeroa ...'} has label \textit{1}. In other words, the first indicator token "gives away" the label. Again, neither of the indicator tokens, \textit{\zeroa} and \textit{\onea}, is predictive of the label if occurring individually.\footnote{The \textit{tic} and \textit{op} shortcuts are implemented so that the special tokens are at most 50 tokens apart.}

Why are contextuality and order worth modeling? Consider the IMDB review example from \citet{ross-etal-2021-explaining} where BERT models learn to rely on numeric ratings present in the input. The learned shortcuts are multi-token -- \textit{`3'} alone is by no means an indicator of a negative review. The order and proximity are important too: \textit{`3'}, \textit{`out'}, \textit{`of'} and \textit{`10'} mentioned far apart or in a different order are not predictive of the negative class. Thus, while other shortcut properties could be proposed, we do believe that the phenomena we model here for lexical shortcuts -- namely, context and order -- are representative of the poor generalization patterns of NLP models. 

\subsection{Creating (Partially) Synthetic Data}

To ensure that the shortcut deterministically indicates the right label, we define shortcuts over tokens absent from the original dataset and introduce them explicitly in the vocabulary\footnote{One could also use existing tokens, provided that there are no counterexamples to the shortcut in the data: e.g., if the shortcut is that \textit{not} signalizes negative label only, there must be no positive inputs mentioning \textit{not}.}. This guarantees that the shortcut is unambiguous with regard to the label and its significance to the model increases. 

Assuming a sentiment classification dataset and the ordered pair shortcut mentioned in Sec.~\ref{sec:shortcuts} (the procedure is analogous for other data-shortcut combinations), we create a synthetic example by (1) randomly sampling an instance from the source data, (2) randomly deciding on the order of the shortcut tokens, (3) inserting these tokens at random positions, obeying the order and (4) setting the label as the shortcut prescribes. This process is illustrated in Fig.~\ref{fig:data} (top left side). In all our experiments the resulting modified datasets are 20\% larger than the source versions. The proportion of the synthetic data was not tuned but picked so that the shortcut data is sufficiently large to be picked by the model but not too large to deteriorate the performance on the unmodified data.

To mitigate the potential problem of making synthetic examples go off-manifold and thus being treated differently by the model as compared with the unmodified examples, for \textit{tic} and \textit{so}, we also inject one of the two tokens from the rule at random into a part of the original data without modifying the label. Thus, for multi-token shortcuts a special token can occur both in examples where it is predictive of the class as well as where it is not (bottom left of Fig.~\ref{fig:data}).

\subsection{Verification Steps}
\label{sec:verification-steps}

The datasets we create are intentionally mixed and consist of the real and synthetic data to approximate real use cases where the model has to extract both simple and complex patterns to perform well. This is different from fully synthetic datasets \cite{yang-etal-2018-mnist-lrp,arras-etal-2019-evaluating} or glass-box DNN models \cite{hao-2020-evaluating} where it is guaranteed that the model uses certain input features but the findings may not be valid for real datasets. Two tests verify that the model indeed uses the shortcut tokens and that they must be most important to the model:
    \begin{enumerate}
        \item The model should achieve close to 100\% accuracy on the fully synthetic test set.\footnote{We also ran experiments on fully synthetic test sets where the shortcut was selected to flip the original label so that there are even more reasons to expect the shortcut tokens to be more important than any other input tokens but got very similar results (see Sec.~\ref{sec:results}).} This would imply that it learned the shortcut and consistently applies it on unseen data (hence the "transparent corner" of the top black box in Fig.~\ref{fig:data}).
        \item The model trained on the original data (the bottom black box in Fig.~\ref{fig:data}) should perform at chance level on the same fully synthetic test set. This would imply that it is indeed the shortcut data and the shortcut rules that are needed to achieve 100\% accuracy. In other words, no other tokens but the shortcut are useful to predict the label in that data.
    \end{enumerate}

\section{Experimental Setup}

We use three text classification datasets and apply the three shortcuts presented above to each of them. Despite all the datasets being binary and of comparable size, there are a few differences which may affect a salience method's performance:

\begin{itemize}
    \item \textbf{SST2} \cite{socher-etal-2013-recursive} is a balanced sentiment classification dataset with short (20 tokens on average) inputs;
    \item \textbf{IMDB} \cite{maas-etal-2011-imdb} is also a balanced sentiment classification dataset with inputs about ten times longer than in SST2; 
    \item \textbf{Toxicity} \cite{wiki-toxicity-data} is a varied length dataset containing toxicity annotations on Wikipedia comments where 9\% of examples are positive (i.e., toxic). Aside from being imbalanced, it differs from the other two in that a text is toxic if it contains a single toxic phrase while for a movie review it is the dominating sentiment which determines the label. 
\end{itemize}

\noindent
In the results section we use the following format to refer to a dataset-shortcut combination: \textit{SST2:tic}, \textit{IMDB:op}, \textit{Toxicity:st}, etc.
\footnote{Modified datasets, models and a demo are available at \url{https://pair-code.github.io/lit/demos/is_eval}.}

\subsection{Models}

We apply the salience methods to explain the predictions of two popular models: a bi-LSTM model \cite{schuster-97-bi-lstm} which uses GloVe embeddings \cite{pennington2014glove}, and BERT \cite{devlin-etal-2019-bert}. Since we only consider binary tasks, the predicted probability of class $c \in \{0, 1\}$ is given by the sigmoid function:

\begin{equation}
    p(c | \mathbf{x}_{1:n}) = \sigma(f_c(\mathbf{x}_{1:n}))
\end{equation}

\noindent where \( f_c(\cdot) \) denotes the model output for class $c$ and \( \mathbf{x}_{1:n} \) is an input of $n$ token embeddings. Both models embed input tokens with a trainable layer so that every \( \mathbf{x}_{i} \) is a continuous $d$-dimensional embedding vector of the $i$-th input token.

The models' accuracy on all the source datasets are presented in Table \ref{tab:model-results}. To verify that the models rely on the introduced shortcuts (Sec.~\ref{sec:verification-steps}), we computed the minimum and mean accuracy on all the nine fully synthetic test sets: these are 99.8 and 99.95 for LSTM and 99.7 and 99.91 for BERT (100\% in most cases). The models trained on the original data (Table \ref{tab:model-results}) all got 50\% accuracy on the same synthetic test sets. 
The close to 100\% performance on the \textit{synthetic} data did not come at the cost of poor performance on the \textit{source} test data: Table~\ref{tab:model-results} reports the mean drop in accuracy averaged over the three shortcut models for each of the dataset and architecture combination. 

\begin{table}[ht]
    \centering
    \begin{tabular}{lccc}
    \toprule
         & SST2 & IMDB & Toxicity \\ %
         \midrule
     LSTM & 87.8 (0.6) & 91.9 (0.0) & 92.5 (0.0) \\ %
     BERT & 93.1 (0.6) & 93.5 (0.2) & 93.2 (0.2) \\ %
     \bottomrule
    \end{tabular}
    \caption{Accuracy on the three \textit{source} (unmodified) test sets of the models trained on the \textit{source} training data. In brackets we report the mean drop in accuracy (on \textit{the same source test sets}) when evaluating the models trained on a shortcut version of the training data.}
    \label{tab:model-results}
\end{table}

\subsection{Salience Methods}

We consider four classes of input salience methods and the Random baseline (\textsc{rand}) to obtain per-token importance weights: Gradient (\textsc{grad*}), Gradient times Input (\textsc{g}x\textsc{i*}), Integrated Gradients (\textsc{ig*}) and \textsc{lime}. 

\subsubsection{Gradient}

\citet{li-etal-2016-visualizing} use gradients as salience weights and compute a score per embedding dimension:
\begin{equation}
    \nabla_{\mathbf{x}_i} f_c(\mathbf{x}_{1:n})\label{eq:grad}
\end{equation}

\noindent To arrive at the per-token score $s(\mathbf{x}_i)$, \citet{li-etal-2016-visualizing}  take the mean absolute value or the $L_1$ norm of the above vector's components. \citet{poerner-etal-2018-evaluating} and \citet{arras-etal-2019-evaluating} use the $L_2$ norm, while \citet{pezeshkpour-etal-2021-combining} use the mean, referencing \citet{atanasova-etal-2020-diagnostic}.

Note that instead of $f_c$ one can compute the gradient of the final layer, that is, in our case the sigmoid function. An argument for starting from the probabilities is that, unlike logits, probabilities contain the information on the relative importance for a particular class. To our knowledge, the effect of using probabilities or logits has not been measured yet. 
In sum, we have six variants of the \textsc{grad} method: \textsc{grad}$_{\{p|l\} \times \{l1|l2|mean\}}$.

\subsubsection{Gradient times Input}

Alternatively, one can compute salience weights by taking the dot product of Eq.~\ref{eq:grad} with the input word embedding $\mathbf{x}_i$ \cite{denil-et-al-2015-extraction} and obtain a salience weight for token $i$:
\begin{equation}
    s(\mathbf{x}_i) = \nabla_{\mathbf{x}_i} f_c(\mathbf{x}_{1:n}) \cdot \mathbf{x}_i
\end{equation}

\noindent
Also here we can compare the probability and the logit versions: \textsc{g}x\textsc{i}$_{\{p|l\}}$.

\subsubsection{Integrated Gradients}

Integrated gradients (\textsc{IG}) \citep{sundararajan-ig-2017} is a gradient-based method which addresses the problem of saturation: gradients may get close to zero for a well-fitted function. IG requires a baseline $\mathbf{b}_{1:n}$ as a way of contrasting the given input with information being absent. A zero vector \cite{mudrakarta-etal-2018-model}, the average embedding or \textsc{unk} or \textsc{[mask]} vectors can serve as baseline vectors in NLP.
For input $i$, we compute:
\begin{equation}
    \frac{1}{m} \!\! \sum_{k=1}^m \!\!\nabla_{\mathbf{x}_i} f_c\!\Big(\!\mathbf{b}_{1:n}\!\!+\! \frac{k}{m} \! (\mathbf{x}_{1:n}\!-\!\mathbf{b}_{1:n} )\!\Big) \!\cdot\!  (\mathbf{x}_i \!-\! \mathbf{b}_i) \!
\end{equation}
\noindent That is, we average over $m$ gradients, with the inputs to $f_c$ being linearly interpolated between the baseline and the original input $\mathbf{x}_{1:n}$ in $m$ steps. We then take the dot product of that averaged gradient with the input embedding $\mathbf{x}_i$ minus the baseline.

In addition to the variable number of steps--small (100) or large (1000)--and the baseline (zero vector, model-specific \textsc{unk} / \textsc{[mask]} or \textsc{pad} / \textsc{[pad]}), also here we can start either from probabilities (i.e., $\sigma$) or logits (i.e., $f$) and arrive at eight different IG configurations: \textsc{ig}$_{\{p|l\} \times \{zero|mask\} \times \{100|1000\}}$. 

\subsubsection{LIME}

\citet{ribeiro-lime} train a linear model to estimate salience of input tokens on a number of perturbations, which are all generated from the given example \( \mathbf{x}_{1:n} \).
A perturbation is an instance where a random subset of tokens in $\mathbf{x}$ is masked out using either \textsc{unk} (LSTM, BERT) or \textsc{[mask]} (BERT) tokens, or dropped completely: \textsc{erase} (LSTM, BERT).
The text model's prediction on these perturbations is the target for the linear model, the masks are the inputs.
Following \citet{ribeiro-lime} we use an exponential kernel with cosine distance and kernel width of 25 as proximity measure of instance and perturbations.
We keep beginning and end-of-sequence tokens unperturbed and experiment with the number of perturbations (100, 1000, 3000).
This results in 6 and 9 configurations for LSTM and BERT, respectively: \textsc{lime}$_{\{unk|mask|erase\} \times \{100|1000|3000\}}$.

\begin{table*}[h]
    \centering
    \setlength{\tabcolsep}{6.5pt}
    \scalebox{0.8}{
    \begin{tabular}{l ccc ccc ccc}
    \toprule
    &  \multicolumn{3}{c}{\textbf{SST2 P}} & \multicolumn{3}{c}{\textbf{IMDB P}} & \multicolumn{3}{c}{\textbf{Toxicity P}} \\
 \cmidrule(lr){2-4}\cmidrule(lr){5-7}\cmidrule(lr){8-10}
     & \textit{st}&\textit{tic}&\textit{op} & \textit{st}&\textit{tic}&\textit{op} & \textit{st}&\textit{tic}&\textit{op}  \\
     \midrule
    LSTM \textsc{g}x\textsc{i}$_{\{p|l\}}$ &\cct{\textbf{1.}}&\cct{\textbf{.76}}&.\textbf{92} &\cct{\textbf{1.}}&\cct{\textbf{.35}}&.\textbf{81} &\textbf{1.}&\textbf{.68}&\textbf{.88}  \\
    \midrule
    BERT \textsc{g}x\textsc{i}$_{\{p|l\}}$ & .29&.58&.31 &.59&.35&.50 &.41&.43&.47 \\
    \bottomrule
    \end{tabular}
    }
    \caption{Precision \textsc{g}x\textsc{i} results across different models and datasets. Here and in the following tables \textbf{P} stands for Precision. Colors and boldface mark the results that are mentioned in the Results section. \textit{st}: single token, \textit{tic}: token in context, \textit{op}: ordered pair. Here we see that  performance of  \textsc{g}x\textsc{i}  varies  across LSTM and BERT, i.e. the LSTM has consistently higher scores on all metrics (in \textbf{bold}). Perfect precision on the single-token shortcut doesn't generalize to strong performance on two-token shortcuts (e.g., \textit{SST2:tic} or \textit{IMDB:tic}, \circlet).
    }
    \label{tab:results:gxi}
\end{table*}

\begin{table*}[h]
    \centering
    \setlength{\tabcolsep}{6.5pt}
    \scalebox{0.75}{
    \begin{tabular}{ll ccc ccc ccc ccc }
    \toprule
   & & \multicolumn{3}{c}{\textbf{SST2 P}} & \multicolumn{3}{c}{\textbf{IMDB P}} & \multicolumn{3}{c}{\textbf{Toxicity P}}  & \multicolumn{3}{c}{\textbf{Toxicity R}}\\
    \cmidrule(lr){3-5}\cmidrule(lr){6-8}\cmidrule(lr){9-11}\cmidrule(lr){12-14}
   & & \textit{st}&\textit{tic}&\textit{op} & \textit{st}&\textit{tic}&\textit{op} & \textit{st}&\textit{tic}&\textit{op}  & \textit{st}&\textit{tic}&\textit{op}  \\
     \midrule
    \parbox[t]{2mm}{\multirow{3}{*}{\rotatebox[origin=c]{90}{LSTM}}} &\textsc{grad}$_{\{p|l\}\times l1}$ & .96&\cct{.50}&\cct{.51} &1.&\cct{.50} & \cct{.52} & .29 & \cct{.53} & \cct{.55} &2&26&13 \\
    &\textsc{grad}$_{\{p|l\}\times l2}$ & .95&\cct{.50}&\cct{.51} &1.&\cct{.50}&\cct{.52} &.37&\cct{.54}&\diagfil{0.68cm}{cb2}{cb1}{.56}  &2&26&\cco{13}\\
    & \textsc{grad}$_{\{p|l\}\mh mean}$ & .28&.23&.28 &.25&.27&.20 &.59&.48&\cco{.60}   &6&31&\cco{21} \\
    \midrule
    \parbox[t]{2mm}{\multirow{3}{*}{\rotatebox[origin=c]{90}{BERT}}} & \textsc{grad}$_{\{p|l\}\times \{l1|l2\}}$ & \textbf{.99}&\textbf{.99}&\textbf{1.} &\textbf{.99}&.87&.96 &\textbf{.99}&\textbf{.99}&\textbf{1. }   &1&2&2 \\
   & \textsc{grad}$_{l\mh mean}$ & \ccth{.41} &\ccth{.44}&\ccth{.42} &\ccth{.41}&\ccth{.38}&\ccth{.41} &\ccth{.41}&\ccth{.45}&\ccth{.44} &48&70&72 \\
   &  \textsc{grad}$_{p\mh mean}$ & \ccth{.43}&\ccth{.45}&\ccth{.42} &\ccth{.39}&\ccth{.34}&\ccth{.41} &\ccth{.44}&\ccth{.46}&\ccth{.44}&43&74&72 \\
    \bottomrule
    \end{tabular}
    }
    \caption{\textsc{grad} Precision across different models and datasets and \textsc{grad} Rank on the Toxcity dataset across models. Here and in the following tables \textbf{R} stands for Rank. \textsc{grad}$_{l2}$ performs very well for BERT (in \textbf{bold}) but not at all so for the LSTM model. The results of \textsc{grad}$_{mean}$ are very poor, ranging between .34 and .46 in precision (\circleth). Rank and precision give complementary information: the precision of \textsc{grad}$_{l2}$ and \textsc{grad}$_{mean}$ is close on \textit{Toxicity:op} (.56 and .60) while the rank of the latter is almost twice as big (13 and 21) (\circleo).}
    \label{tab:results:grad}
\end{table*}

\section{Results}\label{sec:results}

In this section we highlight our main findings. For increased readability where Rank scores support Precision, we omit them in the main paper and instead present them in the Appendix~\ref{sec:rank_res}. All the results reported in the paper are computed from a single model checkpoint and a single run. 

\paragraph{A method's performance varies across model and shortcut types and other dataset properties.} It is apparent that \textsc{g}x\textsc{i} performs quite well for LSTM models but does not work at all for BERT models (Tab.~\ref{tab:results:gxi}). Conversely, \textsc{grad}$_{l2}$ performs very well for BERT but not at all so for LSTM models (Tab.~\ref{tab:results:grad}). Overall, method performance mostly goes down on longer inputs. More interestingly, a strong performance on a simpler shortcut may not persist on a slightly more complex one: \textsc{g}x\textsc{i} has precision of 1.0 on any dataset with the single-token shortcut for LSTM but drops to .76 or even .35 on the same base dataset with a two-token shortcut (e.g., \textit{SST2:tic} or \textit{IMDB:tic}, \circlet in Tab.~\ref{tab:results:gxi}). Thus, even if the model is fixed it cannot be assumed that a certain method works well and would be useful for finding lexical shortcuts learned by the model in general if its evaluation was done on only the single-token shortcut.

\paragraph{\textsc{grad}$_{\{l|p\}\times l*}$ is a good choice for BERT but not LSTM models for finding shortcuts.} For BERT models, \textsc{grad}$_{l2}$ achieves high precision and rank scores across the different datasets and shortcut types, yielding 0.99 or higher on seven out of nine datasets (Tab.~\ref{tab:results:grad} and~\ref{tab:results:grad_r}). The lowest but still comparatively high precision (0.87) is on \textit{IMDB:tic} where the inputs are particularly long. For LSTM models, on six out of nine datasets the precision of the same method is around .5 (\circlet in Tab.~\ref{tab:results:grad}). It does not matter whether probabilities or logits are used and whether L1 or L2 norm is applied. We hypothesize that one reason for the difference in performance between BERT and LSTM is that BERT models have residual connections, making the gradient information less noisy. However, the results of \textsc{grad}$_{mean}$ are very poor, ranging between .3 and .4 in precision (\circleth in Tab.~\ref{tab:results:grad}). 
Note that \textsc{grad}$_{l2}$ is sometimes deemed unsuitable because it is unsigned and only returns positive scores \cite{pezeshkpour-etal-2021-combining}, but our experiments demonstrate that it is the most useful method for finding lexical shortcuts learned by BERT.

\paragraph{Using probabilities instead of logits only changes the results for \textsc{ig}.} For other gradient-based methods it does not seem to make a large difference. (\circlet and \circleth in Tab.~\ref{tab:results:ig} and~\ref{tab:results:ig_r}).

\paragraph{\textsc{ig} performance does not improve much with more steps.} 
Increasing the number of interpolation steps from 100 to 1000 does not result in a significant improvement for LSTM models. Also for BERT, the precision numbers improve only for the \textit{tic} shortcuts and only when probabilities are used (last two rows in Tab.~\ref{tab:results:ig} and~\ref{tab:results:ig_r}).
The similarity of the scores between the  \textsc{g}x\textsc{i} and \textsc{ig} when using the zero baseline (\circleo in Tab.~\ref{tab:results:ig} and~\ref{tab:results:ig_r}) indicates that there is no difference between taking a single or 100(0) steps from the zero baseline.

\begin{table*}[ht]
    \centering
    \setlength{\tabcolsep}{5.5pt}
    \scalebox{0.725}{
    \begin{tabular}{ll ccc ccc ccc}
    \toprule
    & & \multicolumn{3}{c}{\textbf{SST2 P}} & \multicolumn{3}{c}{\textbf{IMDB P}} & \multicolumn{3}{c}{\textbf{Toxicity P}}\\
    \cmidrule(lr){3-5}\cmidrule(lr){6-8}\cmidrule(lr){9-11}
     & & \textit{st}&\textit{tic}&\textit{op} & \textit{st}&\textit{tic}&\textit{op} & \textit{st}&\textit{tic}&\textit{op}  \\
     \midrule
\parbox[t]{2mm}{\multirow{6}{*}{\rotatebox[origin=c]{90}{LSTM}}}  &\textsc{ig}$_{l\mh zero\mh \{100|1000\}}$\footnotemark &         \cct{1.} &             .72 &        \cct{.83} &         \cct{.99} &             .67 &         .80 &         1. &             .95 &         .95  \\
    &\textsc{ig}$_{l\mh unk\mh \{100|1000\}}$ &         1. &             .87 &         .71 &         1. &             .71 &        \ccth{.79} &         1. &             .71 &        \ccth{.78} \\
    &\textsc{ig}$_{l\mh pad \mh \{100|1000\}}$ &     1. &	.77	&.85&	.99&	.67&	.79&	1&	.68&	.67	 \\
    & \textsc{ig}$_{p\mh zero\mh \{100|1000\}}$&         \cct{.93} &             .69 &         \cct{.68} &         \cct{.78} &             .66 &         .76 &         1. &             .93 &         .87  \\
    & \textsc{ig}$_{p\mh unk\mh \{100|1000\}}$&         1. &             .82 &         .77 &         1. &             .70 &        \ccth{.63} &         1. &             .64 &         \ccth{.63}\\
    &\textsc{ig}$_{p\mh pad \mh \{100|1000\}}$ &  .95&	.74&	.78& 	.83&	.67&	.75  \\
     \midrule
    \parbox[t]{2mm}{\multirow{8}{*}{\rotatebox[origin=c]{90}{BERT}}}  &  \textsc{g}x\textsc{i}$_{\{p|l\}}$ & \cco{.29}&\cco{.58}&\cco{.31} &\cco{.59}&\cco{.35}&\cco{.50} &\cco{.41}&\cco{.43}&\cco{.47}  \\
     & \textsc{ig}$_{l\mh zero\mh \{100|1000\}}$&        \cco{.29}&             \cco{.58} &         \cco{.31} &         \cco{.59} &             \cco{.35} &        \cco{.50}&        \cco{.41} &             \cco{.43} &         \cco{.47 } \\
    &\textsc{ig}$_{l\mh mask\mh \{100|1000\}}$ &        \ccth{\textbf{.71}} &          \ccth{\textbf{.58}} &       \ccth{\textbf{.71}}&        \ccth{\textbf{.99}} &            \ccth{\textbf{.62}} &         \ccth{\textbf{.61}} &        \ccth{\textbf{.69}} &            \ccth{\textbf{.50}} &        \ccth{\textbf{.47}}  \\
    &\textsc{ig}$_{l\mh pad \mh \{100|1000\}}$&     .79&   	.27&   	.14&   	.28&   	.47&   	.27&   	.36&   	.46&   	.18 \\
    
   & \textsc{ig}$_{p\mh zero\mh \{100|1000\}}$  &         .\cctf{29} &             \cctf{.58} &         \cctf{.31} &        \cctf{ .59} &             \cctf{.35} &         \cctf{.50} &         \cctf{.41} &            \cctf{ .43} &         \cctf{.47}\\
    &\textsc{ig}$_{p\mh  mask\mh 100}$ &       \diagfil{0.55cm}{cb4}{cb3}{.48} &              \diagfil{0.55cm}{cb4}{cb3}{.37} &          \diagfil{0.55cm}{cb4}{cb3}{.56} &          \diagfil{0.60cm}{cb4}{cb3}{.80} &               \diagfil{0.55cm}{cb4}{cb3}{.34} &           \diagfil{0.55cm}{cb4}{cb3}{.48} &          \diagfil{0.60cm}{cb4}{cb3}{.27} &               \diagfil{0.60cm}{cb4}{cb3}{.27} &           \diagfil{0.55cm}{cb4}{cb3}{.29}\\
    &\textsc{ig}$_{p\mh  mask\mh 1000}$ &   .48 &             .48 &         .56 &         .80 &             .47 &         .48 &         .28 &             .29 &         .29\\
        &\textsc{ig}$_{p\mh pad \mh \{100|1000\}}$ &    \cctf{.81} &	\cctf{.18}&	\cctf{.1}&	\cctf{.21}&	\cctf{.31}&	\cctf{.14}&	\cctf{.16}&	\cctf{.37}&	\cctf{.12}	\\
    \bottomrule
    \end{tabular}
    }
    \caption{IG Precision across different models and datasets. Using probabilities instead of logits changes the results for \textsc{ig} (\circlet and \circleth). Number of steps doesn't affect the \textsc{ig} performance, but the choice of the baseline is important for \textsc{ig} when using BERT (\circlef). Using the \textsc{[mask]} baseline (with logits) resulted in an improvement in the  scores (\textbf{bold}). Finally, \circleo rows tell us that the difference between  \textsc{g}x\textsc{i} and \textsc{ig}$_{p\mh zero\mh \{100|1000\}}$  is minimal and there is no difference between taking a single or 100(0) steps from the zero baseline.
    }
    \label{tab:results:ig}
\end{table*}

\begin{table*}[ht]
    \centering
    \setlength{\tabcolsep}{6.5pt}
    \scalebox{0.79}{
    \begin{tabular}{ll ccc ccc ccc}
    \toprule
    & &\multicolumn{3}{c}{\textbf{SST2 P}} & \multicolumn{3}{c}{\textbf{IMDB P}} & \multicolumn{3}{c}{\textbf{Toxicity P}} \\
     \cmidrule(lr){3-5}\cmidrule(lr){6-8}\cmidrule(lr){9-11}
     & &\textit{st}&\textit{tic}&\textit{op} & \textit{st}&\textit{tic}&\textit{op} & \textit{st}&\textit{tic}&\textit{op}  \\
     \midrule
    \parbox[t]{2mm}{\multirow{3}{*}{\rotatebox[origin=c]{90}{LSTM}}} &\textsc{lime}$_{unk-100}$ &  .98 &             .80 &         .83 &         .92 &             .50 &         .62 &         .93 &             .58 &         .59  \\
    &\textsc{lime}$_{unk-1000}$ & 1. &             .83 &         .85 &         .99 &             .66 &         .78 &         1. &             .84 &         .66  \\
    &\textsc{lime}$_{unk-3000}$ & 1. &             .84 &         .85 &         1. &             .66 &         .80 &         1. &             .85 &         .66  \\
     \midrule
   \parbox[t]{2mm}{\multirow{4}{*}{\rotatebox[origin=c]{90}{BERT}}} &\textsc{lime}$_{unk-100}$ & .89 &             .80 &         .71 &         .91 &             .62 &         .44 &         .67 &             .54 &         .51 \\
    &\textsc{lime}$_{unk-1000}$ & .97 &             .87 &         .77 &         .99 &             .75 &         .70 &         .98 &             .58 &         .75  \\
    &\textsc{lime}$_{unk-3000}$ & \cct{.98} &             \cct{.88} &         \cct{.77} &         \cct{.99} &             \cct{.77} &         \cct{.71} &         \cct{1.} &             \cct{.59} &         \cct{.78} \\
    & \textsc{lime}$_{mask-3000}$ & \cct{.98} &             \cct{.62} &         \cct{.78} &         \cct{.93} &             \cct{.76} &         \cct{.67} &         \cct{.99} &             \cct{.58} &         \cct{.70}  \\
    \bottomrule
    \end{tabular}
    }
    \caption{LIME Precision across different models and datasets. LIME benefits from 1000 over 100 perturbations, especially for longer inputs and/or shortcuts. We found that the increase from 1000 to 3000 perturbations leads to little precision improvements for the input lengths in our datasets. Using \textsc{unk} for masking leads to better results than \textsc{[mask]} in several configurations (\circlet).
    }
    \label{tab:results:lime}
\end{table*}

\paragraph{Choice of baseline is important for \textsc{ig} when using BERT.}
For the most part using the \textsc{[mask]} baseline (with logits) resulted in an improvement in the  scores (\textbf{bold} rows in Tab.~\ref{tab:results:ig} and ~\ref{tab:results:ig_r}). Still, even with the best performing configuration of \textsc{ig} the results are much worse than \textsc{grad}$_{l \mh l2}$.

\paragraph{Number of perturbations as well as masking token matter for \textsc{lime}.}
LIME benefits from 1000 over 100 perturbations, especially for longer inputs and/or shortcuts.
We found that the increase from 1000 to 3000 perturbations leads to little precision improvements for the input lengths in our datasets.
Using \textsc{unk} for masking leads to better results than \textsc{[mask]} in almost all configurations (\circlet in Tab.~\ref{tab:results:lime} and~\ref{tab:results:lime_r}).
We hypothesize this is due to two reasons:
(i) The \textsc{[mask]} token is not used during fine-tuning on the task data.
(ii) The \textsc{unk} token, however, is finetuned (due to unknown tokens and as special token in word dropout).
Erasing tokens leads, on average, to worse precision results than masking, for all number of perturbations. Tables \ref{tab:results:lstm} and \ref{tab:results:bert} in Appendix~\ref{sec:full_res} present the results for all the models, shortcut types and source datasets in terms of precision and rank and you can observe this phenomenon there.

\paragraph{Rank and precision give complementary information.} 
For example, the precision of \textsc{grad}$_{l2}$ and \textsc{grad}$_{mean}$ is close on \textit{Toxicity:op} (.56 and .60) while the rank of the latter is almost twice as big (13 and 21) (\circleo in Tab.~\ref{tab:results:grad}). Lower rank with comparable precision means that the method consistently puts one of the shortcut tokens on the top but buries the other token deep in the ranking.

\footnotetext{The score differences between 100 and 1000 steps for this and the following methods is within 3\%.}

\section{Related Work}

Research on input salience methods for text classification is prolific and diverse in terms of the definitions used \cite{camburu2020struggles}, applications \cite{feng-graber-2019-what}, desiderata \cite{sundararajan-2017-ig}, etc. The importance of getting faithful salience explanations has been recognized early on \cite{bach-etal-2015,kindermans-etal-2017} and there exist formal definitions of explanation fidelity \cite{yeh-etal-2019-fidelity}.
However, these have not been connected to model debugging where it is the top of a salience ranking that matters most. 
In the vision domain, our work is closest to \citet{adebayo-etal-2020-debugging,adebayo-etal-2022-post}, who also explore the debugging scenario with salience maps, \citet{yang-kim-2019}, who use synthetic data to obtain the ground truth for pixel importance, and \citet{hooker-etal-2019-benchmark}, who contrast the performance of the same model trained on original and modified data when evaluating feature importance. 

As pointed out in Introduction, in NLP faithfulness evaluation has often been grounded in strong assumptions \cite{poerner-etal-2018-evaluating,deyoung-etal-2020-eraser,atanasova-etal-2020-diagnostic,ding-koehn-2021-evaluating} or by analyzing models substantially different from the ones normally used \cite{arras-etal-2019-evaluating,hao-2020-evaluating}. 
An exception to this trend is the work by \citet{sippy-2020-staining} who also modify source data but, unlike us, consider MLP as the only DNN model, do not evaluate any gradient-based methods and analyze single token shortcuts only without strong guarantees of them actually being the most important clues for the model. Also \citet{zhou-et-al-2021-do-feature-attribution} analyze DNN models on intentionally corrupted data: they primarily focus on vision but also run an experiment analyzing how faithfully the attention mechanism points at the words known to correlate with the label.
Finally, \citet{madsen-et-al-2021-evaluating}, following \citet{hooker-2018-roar}, iteratively remove tokens to evaluate faithfulness of salience methods for LSTM models and conclude, similar to us, %
that performance is task-dependent.

Concurrently with our work, \citet{idahl-etal-2021-benchmarking} argue for faithfulness evaluation on synthetic data for model debugging but do not report experimental results. 
Similarly to them and also concurrently with our work, \citet{pezeshkpour-etal-2021-combining} go further and combine data and input attribution methods to discover data artifacts. %
However, citing prior work, they use \textsc{grad}$_{l\mh mean}$ and \textsc{ig}$_{l\mh mean}$ which, as we have shown, are sub-optimal configurations for BERT models. This explains the very poor accuracy of 12-13\% (in our terms: precision@1) that they observed when discovering single-token shortcuts in SST2. Finally, as our experiments demonstrate, the single-token shortcut is insufficient to assess whether a method would be useful for more complex shortcuts.

\section{Conclusions}

We have argued for evaluating input salience methods with respect to how helpful they would be for discovering shortcuts that are learned by the model. This seems to be a clear use case from the model developer perspective. To achieve this, we proposed a protocol for method evaluation and applied it to three variants of lexical shortcuts (single token, token in context, and ordered pair) which are a proxy for shortcut heuristics that occur in common NLP tasks and which are particularly suitable for being discovered with input salience methods. By comparing the performance across different datasets, shortcut types and models (LSTM-based and BERT-based), we demonstrated that a strong performance for one setup may not hold for a different model or a more complex shortcut. Finally, we pointed out that some method configurations assumed to be reliable in recent work, for example integrated gradients, may give very poor results for NLP models, and that the details of how the methods are used can matter a lot, such as how a gradient vector is reduced into a scalar. Our results demonstrate that whenever one uses BERT and is interested if a simple token combination could determine the label, one should prefer Grad-L2 over more complex methods.

\section{Limitations}
In this paper we proposed a protocol that can be used for evaluating input salience methods. We limited ourselves to the most popular salience methods, and left others out of scope. 
In particular, it would be of interest to evaluate the most recent salience methods, like \citet{chen-etal-2020-generating,sikdar-etal-2021-integrated}, which were developed to take feature interactions into account.
We also limited this work to the task of English (binary) text classification. Furthermore, we focus on a representative set of shortcuts, but different shortcuts might result in different outcomes.
Finally, we limited ourselves to LSTM and BERT based models. Results with different neural components or with models of a different size and/or depth may be different. However, the protocol that we proposed can still be used in those cases. We also note that input salience is only one kind of explanation, and a limited one: it does not reveal the logic of the model, nor does it reveal interactions between input features. It is hardly possible to fully understand why a deep non-linear
neural model produced a certain prediction by \emph{only} looking at input salience scores.

\paragraph{Acknowledgements} We would like to thank Ian Tenney, Alon Jacovi and the anonymous reviewers for the many helpful suggestions. 

\bibliographystyle{acl_natbib}
\bibliography{anthology,custom}
\newpage

\clearpage
\appendix

\section{Appendix}
\label{sec:appendix}
\subsection{Architecture and training details}

\paragraph{LSTM} For all datasets we use the same LSTM model consisting of an embedding layer that is initialized with a pretrained GloVe embedding, a single bidirectional LSTM layer and an attention classifier layer. The attention classifier consists of a keys-only attention layer~\cite{bahdanau2014neural, jain-wallace-2019-attention} followed by a single layer MLP. The embedding size is $300$ and the word dropout rate is $0.1$, the LSTM and the classifier hidden sizes are set to $256$. The output size of the classifier is set to $1$. During training we use the same dropout rate of $0.5$ in all layers. We train this model with an SGD optimizer with a learning rate of $0.03$, momentum $0.9$ and a weight decay of $5\mathrm{e}{-6}$. We trained our models at most for $35000$ steps, however we enabled early stopping if after $10000$ steps  we didn't observe scores improvement on the validation set. For SST2 and IMDB we used a batch size of $64$ and for the Toxicity we used the size of $32$. With these hyper-parameters the LSTM models contain $5468048$ (5.4M) parameters.

\paragraph{BERT}
For all datasets we use BERT Base model: 12 layers, 12 heads and a hidden size of 768. We load the publicly available pretrained uncased checkpoint before finetuning on our data. During training we use the same dropout rate of $0.5$. We chose the ADAM optimizer, with a learning rate of $2\mathrm{e}{-5}$ and a weight decay of $5\mathrm{e}{-6}$ following the best practices of finetuning BERT~\cite{devlin-etal-2019-bert}. For Toxicity we set the learning rate to $1\mathrm{e}{-5}$, the rest of the parameters are the same. The maximum sequence length in SST2 we set to 100, and for IMDB and Toxicity we set it to 500. We follow the same early stopping configuration as in LSTM. We use a batch size of 16 everywhere. With these hyper-parameters the BERT (Base) models contain $109483009$ (109M) parameters.

\subsection{Budget}
For our experiments we used a total of 35 hours on TPUv2 (4-core) accelerators.

\subsection{Methods implementation details}
\paragraph{Integrated gradients} For the non-zero IG baseline, we take the sequence of embedded inputs, keep the embeddings of the special tokens (e.g. CLS and SEP) the same, and replace the other embedded inputs with the embedded baseline token (e.g., MASK or UNK).

\subsection{Rank results}
\label{sec:rank_res}
Please refer to Tables~\ref{tab:results:gxi_r},~\ref{tab:results:ig_r},~\ref{tab:results:grad_r},~\ref{tab:results:lime_r}  for the Rank results that complete the story in the paper.
\begin{table*}[h]
    \centering
    \setlength{\tabcolsep}{6.5pt}
    \scalebox{0.8}{
    \begin{tabular}{l ccc ccc ccc}
    \toprule
    &  \multicolumn{3}{c}{\textbf{SST2}} & \multicolumn{3}{c}{\textbf{IMDB}} & \multicolumn{3}{c}{\textbf{Toxicity}} \\
    \cmidrule(lr){2-4}\cmidrule(lr){5-7}\cmidrule(lr){8-10}\
     & \textit{st}&\textit{tic}&\textit{op} & \textit{st}&\textit{tic}&\textit{op} & \textit{st}&\textit{tic}&\textit{op} \\
     \midrule
    LSTM & \textbf{1}&\textbf{8}&\textbf{3} &\textbf{1}&\textbf{212}&\textbf{24} &\textbf{1}&\textbf{16}&\textbf{5} \\
    \midrule
    BERT   &17&17&24 &106&214&189 &33&88&69 \\
    \bottomrule
    \end{tabular}
    }
    
    \caption{\textsc{g}x\textsc{i} Rank results across different models and datasets. Here we see that  performance of  \textsc{g}x\textsc{i}  varies  across LSTM and BERT, i.e. LSTM has consistently higher scores on all metrics (in \textbf{bold}).
    }
    \label{tab:results:gxi_r}
\end{table*}

\begin{table*}[h]
    \centering
    \setlength{\tabcolsep}{6.5pt}
    \scalebox{0.8}{
    \begin{tabular}{l ccc ccc ccc}
    \toprule
   &  \multicolumn{3}{c}{\textbf{SST2 R}} & \multicolumn{3}{c}{\textbf{IMDB R}} & \multicolumn{3}{c}{\textbf{Toxicity R}}\\
    \cmidrule(lr){2-4}\cmidrule(lr){5-7}\cmidrule(lr){8-10}
   &  \textit{st}&\textit{tic}&\textit{op} & \textit{st}&\textit{tic}&\textit{op} & \textit{st}&\textit{tic}&\textit{op} \\
     \midrule
    \parbox[t]{10mm}{\multirow{3}{*}{LSTM}}
     &1&11&15 &1&22&51 &2&26&13 \\
     &1&11&15 &1&22&50 &2&26&\cco{13}\\
     &7&21&20 &152&123&153 &6&31&\cco{21} \\
    \midrule
    \parbox[t]{10mm}{\multirow{3}{*}{BERT}}
    &1&2&2 &1&2&2 &1&2&2 \\
    &12&21&21 &132&203&206 &48&70&72 \\
    &13&21&21 &134&204&204 &43&74&72 \\
    \bottomrule
    \end{tabular}
    }
    \caption{\textsc{grad} Rank across different models and datasets. }
    \label{tab:results:grad_r}
\end{table*}

\begin{table*}[ht]
    \centering
    \setlength{\tabcolsep}{5.5pt}
    \scalebox{0.725}{
    \begin{tabular}{ll ccc ccc ccc}
    \toprule
    & &  \multicolumn{3}{c}{\textbf{SST2 R}} & \multicolumn{3}{c}{\textbf{IMDB R}} & \multicolumn{3}{c}{\textbf{Toxicity R}}\\
    \cmidrule(lr){3-5}\cmidrule(lr){6-8}\cmidrule(lr){9-11}
     & &  \textit{st}&\textit{tic}&\textit{op} & \textit{st}&\textit{tic}&\textit{op} & \textit{st}&\textit{tic}&\textit{op} \\
     \midrule
     \parbox[t]{10mm}{\multirow{6}{*}{LSTM}}  &\textsc{ig}$_{l\mh zero\mh \{100|1000\}}$ &     1 &               11 &            6 &            1 &              118 &           97 &            1 &                3 &            3 \\
    &\textsc{ig}$_{l\mh unk\mh \{100|1000\}}$ &           1 &                6 &           12 &            1 &              112 &          \ccth{94} &            1 &               \ccth{29} &          \ccth{20} \\
    &\textsc{ig}$_{l\mh pad \mh \{100|1000\}}$ &  	1&	10&	6&	1&	126&	100&	1&	26&	20							 \\
    & \textsc{ig}$_{p\mh zero\mh \{100|1000\}}$&          1 &               11 &            8 &            3 &              118 &           97 &            1 &                3 &            9 \\
    & \textsc{ig}$_{p\mh unk\mh \{100|1000\}}$&           1 &                8 &           10 &            1 &              117 &          \ccth{117} &            1 &               \ccth{39}&           \ccth{36} \\
    &\textsc{ig}$_{p\mh pad \mh \{100|1000\}}$& 	1&	10&	7&  2&	126&	105&	1&	44&	32	 \\
     \midrule
    \parbox[t]{10mm}{\multirow{8}{*}{BERT}}  &   \textsc{g}x\textsc{i}$_{\{p|l\}}$ & \cco{17}&\cco{17}&\cco{24} &\cco{106}&\cco{214}&\cco{189} &\cco{33}&\cco{88}&\cco{69} \\
     & \textsc{ig}$_{l\mh zero\mh \{100|1000\}}$ &             \cco{16}&               \cco{17}&           \cco{24} &          \cco{105} &              \cco{213} &          \cco{190} &           \cco{33} &               \cco{88} &           \cco{69} \\
    &\textsc{ig}$_{l\mh mask\mh \{100|1000\}}$ &            \ccth{\textbf{2}} &              \ccth{\textbf{13}} &           \ccth{\textbf{6}}&           \ccth{\textbf{1}} &            \ccth{\textbf{110}} &          \ccth{\textbf{82} }&            \ccth{\textbf{4}} &               \ccth{\textbf{63}} &           \ccth{\textbf{40}} \\
    &\textsc{ig}$_{l\mh pad \mh \{100|1000\}}$ &   	4&   	18	&   14&   	120&   	243&   	141&   	3&   	92&   	31								 \\
    
   & \textsc{ig}$_{p\mh zero\mh \{100|1000\}}$ &             \cctf{ 16} &              \cctf{ 17} &           \cctf{24} &          \cctf{105} &             \cctf{213} &         \cctf{ 190 }&           \cctf{33} &               \cctf{88} &           \cctf{69} \\
    &\textsc{ig}$_{p\mh  mask\mh 100}$ &          \diagfil{0.55cm}{cb4}{cb3}{5} &                \diagfil{0.55cm}{cb4}{cb3}{14} &             \diagfil{0.55cm}{cb4}{cb3}{8} &             \diagfil{0.60cm}{cb4}{cb3}{17} &               \diagfil{0.60cm}{cb4}{cb3}{15} &             \diagfil{0.70cm}{cb4}{cb3}{93} &             \diagfil{0.55cm}{cb4}{cb3}{31} &                 \diagfil{0.55cm}{cb4}{cb3}{64} &             \diagfil{0.55cm}{cb4}{cb3}{50} \\
    &\textsc{ig}$_{p\mh  mask\mh 1000}$ &         5 &               13 &            9 &           10 &              106 &           95 &           31 &               66 &           50 \\
        &\textsc{ig}$_{p\mh pad \mh \{100|1000\}}$ & 	\cctf{3}&	\cctf{18}&	\cctf{15}&	\cctf{114}&	\cctf{219}&	\cctf{153}&	\cctf{11}&	\cctf{83}	&\cctf{47}	\\
    \bottomrule
    \end{tabular}
    }
    \caption{IG Rank across different models and datasets.   Similarly to the precision results from Tab.~\ref{tab:results:ig}  we see that using probabilities instead of logits changes the results for \textsc{ig} (\circleth). We also observe that number of steps doesn't affect the \textsc{ig} performance, but the choice of baseline is important for \textsc{ig} when using BERT (\circlef). Finally \circleo rows tell us the difference between  \textsc{g}x\textsc{i} and \textsc{ig}$_{p\mh zero\mh \{100|1000\}}$  is minimal and there is no difference between taking a single or 100(0) steps from the zero baseline.
    }
    \label{tab:results:ig_r}
\end{table*}

\begin{table*}[ht]
    \centering
    \setlength{\tabcolsep}{6.5pt}
    \scalebox{0.79}{
    \begin{tabular}{ll ccc ccc ccc }
    \toprule
    &  & \multicolumn{3}{c}{\textbf{SST2 R}} & \multicolumn{3}{c}{\textbf{IMDB R}} & \multicolumn{3}{c}{\textbf{Toxicity R}}\\
     \cmidrule(lr){3-5}\cmidrule(lr){6-8}\cmidrule(lr){9-11}
     & &\textit{st}&\textit{tic}&\textit{op} & \textit{st}&\textit{tic}&\textit{op} & \textit{st}&\textit{tic}&\textit{op}\\
     \midrule
    \parbox[t]{2mm}{\multirow{3}{*}{\rotatebox[origin=c]{90}{LSTM}}} &\textsc{lime}$_{unk-100}$ &    1 &             7 &         4 &         2 &            99 &        67 &         1 &            13 &        34 \\
    &\textsc{lime}$_{unk-1000}$ &    1 &             6 &         4 &         1&            83 &        45 &         1 &             7 &        34 \\
    &\textsc{lime}$_{unk-3000}$ &       1 &                6 &            4 &            1 &               82 &           40 &            1 &                8 &           33 \\
     \midrule
   \parbox[t]{2mm}{\multirow{4}{*}{\rotatebox[origin=c]{90}{BERT}}} &\textsc{lime}$_{unk-100}$ &     1 &             8 &         7 &         3 &            94 &        67 &         6 &            82 &        28 \\
    &\textsc{lime}$_{unk-1000}$ &   1 &             7 &         6 &         2 &            96 &        57 &         1 &            78 &        23 \\
    &\textsc{lime}$_{unk-3000}$ &     \cct{1} &                \cct{7} &            \cct{5} &            \cct{2} &               \cct{97} &           \cct{56} &            \cct{1} &               \cct{77} &           \cct{21} \\
    & \textsc{lime}$_{mask-3000}$ &    \cct{1} &            \cct{13} &         \cct{5} &        \cct{13} &           \cct{105} &        \cct{52} &         \cct{1} &            \cct{80} &        \cct{12} \\
    \bottomrule
    \end{tabular}
    }
    \caption{LIME Rank across different models and datasets.  Similarly to the precision results from Tab.~\ref{tab:results:lime} we see that LIME benefits from 1000 over 100 perturbations, especially for longer inputs and/or shortcuts. We found that the increase from 1000 to 3000 perturbations leads to little precision improvements for the input lengths in our datasets. Using \textsc{unk} for masking leads to better results than \textsc{[mask]} in almost all configurations (\circlet).
    }
    \label{tab:results:lime_r}
\end{table*}

\subsection{Full Results}
\label{sec:full_res}
Table~\ref{tab:results:lstm} and Table~\ref{tab:results:bert} list the full results for LSTM and BERT respectively.
\begin{table*}[]
    \centering
    \setlength{\tabcolsep}{6.5pt}
    \scalebox{0.8}{
    \begin{tabular}{l ccc ccc ccc ccc ccc ccc}
    \toprule
    &  \multicolumn{3}{c}{\textbf{SST2}} & \multicolumn{3}{c}{\textbf{IMDB}} & \multicolumn{3}{c}{\textbf{Toxicity}} & \multicolumn{3}{c}{\textbf{SST2}} & \multicolumn{3}{c}{\textbf{IMDB}} & \multicolumn{3}{c}{\textbf{Toxicity}}\\
 & \multicolumn{3}{c}{Precision} & \multicolumn{3}{c}{Precision} & \multicolumn{3}{c}{Precision} & \multicolumn{3}{c}{Rank} & \multicolumn{3}{c}{Rank} & \multicolumn{3}{c}{Rank} \\   
   \cmidrule(lr){2-4}\cmidrule(lr){5-7}\cmidrule(lr){8-10}\cmidrule(lr){11-13}\cmidrule(lr){14-16}\cmidrule(lr){17-19}
     & \textit{st}&\textit{tic}&\textit{op} & \textit{st}&\textit{tic}&\textit{op} & \textit{st}&\textit{tic}&\textit{op} & \textit{st}&\textit{tic}&\textit{op} & \textit{st}&\textit{tic}&\textit{op} & \textit{st}&\textit{tic}&\textit{op} \\
     \midrule
    \textsc{random} & .06&.1&.1 &.0&.01&.01 &.03&.07&.06     &11&16&17 &120&161&162 &27&36&36 \\
    \textsc{grad}$_{\{p|l\}\times l1}$ & .96&.5&.51 &1.&.5&.52 &.29&.53&.55     &1&11&15 &1&22&51 &2&26&13 \\
    \textsc{grad}$_{\{p|l\}\times l2}$ & .95&.5&.51 &1.&.5&.52 &.37&.54&.56     &1&11&15 &1&22&50 &2&26&13 \\
    \textsc{grad}$_{\{p|l\}\mh mean}$ & .28&.23&.28 &.25&.27&.2 &.59&.48&.6     &7&21&20 &152&123&153 &6&31&21 \\
    \textsc{g}x\textsc{i}$_{\{p|l\}}$ & 1.&.76&.92 &1.&.35&.81 &1.&.68&.88     &1&8&3 &1&212&24 &1&16&5 \\
     \textsc{ig}$_{l\mh zero\mh 100}$&         1 &             .72 &         .83 &         .99 &             .67 &         .80 &         1 &             .95 &         .95 &            1 &               11 &            6 &            1 &              118 &           97 &            1 &                3 &            3 \\
     \textsc{ig}$_{l\mh zero\mh 1000}$&         1 &             .72 &         .83 &         1 &             .67 &         .79 &         1 &             .95 &         .95 &            1 &               11 &            6 &            1 &              118 &           95 &            1 &                3 &            3 \\
    \textsc{ig}$_{l\mh unk\mh 100}$ &         1 &             .87 &         .71 &         1 &             .71 &         .79 &         1 &             .71 &         .78 &            1 &                6 &           12 &            1 &              112 &           94 &            1 &               29 &           20 \\
     \textsc{ig}$_{l\mh unk\mh 1000}$&         1 &             .88 &         .71 &         1 &             .71 &         .81 &         1 &             .71 &         .78 &            1 &                6 &           12 &            1 &              113 &           89 &            1 &               30 &           20 \\
     \textsc{ig}$_{p\mh zero\mh 100}$&         .93 &             .69 &         .68 &         .78 &             .66 &         .76 &         1 &             .93 &         .87 &            1 &               11 &            8 &            3 &              118 &           97 &            1 &                3 &            9 \\
     \textsc{ig}$_{p\mh zero\mh 1000}$&         .93 &             .69 &         .68 &         .75 &             .66 &         .77 &         1 &             .93 &         .87 &            1 &               11 &            8 &            3 &              118 &           95 &            1 &                3 &            9 \\
     \textsc{ig}$_{p\mh unk\mh 100}$&         1 &             .82 &         .77 &         1 &             .70 &         .63 &         1 &             .64 &         .63 &            1 &                8 &           10 &            1 &              117 &          117 &            1 &               39 &           36 \\
     \textsc{ig}$_{p\mh unk\mh 1000}$&         1 &             .82 &         .77 &         1 &             .70 &         .65 &         1 &             .64 &         .62 &            1 &                8 &           10 &            1 &              117 &          111 &            1 &               40 &           37 \\

    \textsc{lime}$_{unk-100}$ &  .98 &             .80 &         .83 &         .92 &             .50 &         .62 &         .93 &             .58 &         .59 &         1 &             7 &         4 &         2 &            99 &        67 &         1 &            13 &        34 \\
    \textsc{lime}$_{unk-1000}$ & 1. &             .83 &         .85 &         .99 &             .66 &         .78 &         1. &             .84 &         .66 &         1 &             6 &         4 &         1&            83 &        45 &         1 &             7 &        34 \\
    \textsc{lime}$_{unk-3000}$ & 1. &             .84 &         .85 &         1. &             .66 &         .80 &         1. &             .85 &         .66 &            1 &                6 &            4 &            1 &               82 &           40 &            1 &                8 &           33 \\
    \textsc{lime}$_{erase-100}$ & .93 &             .68 &         .74 &         .77 &             .50 &         .60 &         .97 &             .69 &         .44 &            1 &                8 &            9 &            7 &              120 &           75 &            1 &                9 &           32 \\
    \textsc{lime}$_{erase-1000}$ & .99 &             .72 &         .77 &         .97 &             .72 &         .76 &         1. &             .94 &         .68 &            1 &                7 &            8 &            1 &               83 &           55 &            1 &                2 &           33 \\
    \textsc{lime}$_{erase-3000}$ & 1. &             .72 &         .77 &         .99 &             .73 &         .77 &         1. &             .96 &         .70 &            1 &                7 &            8 &            1 &               76 &           50 &            1 &                2 &           33 \\
    
    \bottomrule
    \end{tabular}
    }
    \caption{Precision and rank of all the method configurations across the datasets and shortcut types for LSTM. \textit{st}: single token, \textit{tic}: token in context, \textit{op}: ordered pair.
    }
    \label{tab:results:lstm}
\end{table*}

\begin{table*}[]
    \centering
    \setlength{\tabcolsep}{6.5pt}
    \scalebox{0.8}{
    \begin{tabular}{l ccc ccc ccc ccc ccc ccc}
    \toprule
    &  \multicolumn{3}{c}{\textbf{SST2}} & \multicolumn{3}{c}{\textbf{IMDB}} & \multicolumn{3}{c}{\textbf{Toxicity}} & \multicolumn{3}{c}{\textbf{SST2}} & \multicolumn{3}{c}{\textbf{IMDB}} & \multicolumn{3}{c}{\textbf{Toxicity}}\\
 & \multicolumn{3}{c}{Precision} & \multicolumn{3}{c}{Precision} & \multicolumn{3}{c}{Precision} & \multicolumn{3}{c}{Rank} & \multicolumn{3}{c}{Rank} & \multicolumn{3}{c}{Rank} \\
  \cmidrule(lr){2-4}\cmidrule(lr){5-7}\cmidrule(lr){8-10}\cmidrule(lr){11-13}\cmidrule(lr){14-16}\cmidrule(lr){17-19}
     & \textit{st}&\textit{tic}&\textit{op} & \textit{st}&\textit{tic}&\textit{op} & \textit{st}&\textit{tic}&\textit{op} & \textit{st}&\textit{tic}&\textit{op} & \textit{st}&\textit{tic}&\textit{op} & \textit{st}&\textit{tic}&\textit{op} \\
     \midrule
    \textsc{grad}$_{\{p|l\}\times \{l1|l2\}}$ & .99&.99&1. &.99&.87&.96 &.99&.99&1.     &1&2&2 &1&2&2 &1&2&2 \\
    \textsc{grad}$_{l\mh mean}$ & .41&.44&.42 &.41&.38&.41 &.41&.45&.44     &12&21&21 &132&203&206 &48&70&72 \\
    \textsc{grad}$_{p\mh mean}$ & .43&.45&.42 &.39&.34&.41 &.44&.46&.44     &13&21&21 &134&204&204 &43&74&72 \\
    \textsc{g}x\textsc{i}$_{\{p|l\}}$ & .29&.58&.31 &.59&.35&.50 &.41&.43&.47     &17&17&24 &106&214&189 &33&88&69 \\
\textsc{ig}$_{l\mh zero\mh \{100 | 1000\}}$ &         .29 &             .58 &         .31 &         .59 &             .35 &         .50 &         .41 &             .43 &         .47 &           16 &               17 &           24 &          105 &              213 &          190 &           33 &               88 &           69 \\
    \textsc{ig}$_{l\mh mask\mh 100}$ &         .71 &             .58 &         .71 &         .99 &             .62 &         .61 &         .69 &             .50 &         .47 &            2 &               13 &            6 &            1 &              110 &           82 &            4 &               63 &           40 \\
\textsc{ig}$_{l\mh  mask\mh 1000}$ &         .71 &             .58 &         .71 &         .99 &             .64 &         .61 &         .70 &             .50 &         .47 &            2 &               14 &            5 &            1 &              109 &           83 &            4 &               65 &           41 \\
\textsc{ig}$_{p\mh zero\mh \{100|1000\}}$ &         .29 &             .58 &         .31 &         .59 &             .35 &         .50 &         .41 &             .43 &         .47 &           16 &               17 &           24 &          105 &              213 &          190 &           33 &               88 &           69 \\
\textsc{ig}$_{p\mh  mask\mh 100}$ &         .48 &             .37 &         .56 &         .80 &             .34 &         .48 &         .27 &             .27 &         .29 &            5 &               14 &            8 &           17 &              115 &           93 &           31 &               64 &           50 \\
\textsc{ig}$_{p\mh  mask\mh 1000}$ &         .48 &             .48 &         .56 &         .80 &             .47 &         .48 &         .28 &             .29 &         .29 &            5 &               13 &            9 &           10 &              106 &           95 &           31 &               66 &           50 \\
    \textsc{lime}$_{unk-100}$ & .89 &             .80 &         .71 &         .91 &             .62 &         .44 &         .67 &             .54 &         .51 &         1 &             8 &         7 &         3 &            94 &        67 &         6 &            82 &        28 \\
    \textsc{lime}$_{unk-1000}$ & .97 &             .87 &         .77 &         .99 &             .75 &         .70 &         .98 &             .58 &         .75 &         1 &             7 &         6 &         2 &            96 &        57 &         1 &            78 &        23 \\
    \textsc{lime}$_{unk-3000}$ & .98 &             .88 &         .77 &         .99 &             .77 &         .71 &         1. &             .59 &         .78 &            1 &                7 &            5 &            2 &               97 &           56 &            1 &               77 &           21 \\
    \textsc{lime}$_{mask-3000}$ & .98 &             .62 &         .78 &         .93 &             .76 &         .67 &         .99 &             .58 &         .70 &         1 &            13 &         5 &        13 &           105 &        52 &         1 &            80 &        12 \\
    \textsc{lime}$_{erase-100}$ & .83 &             .61 &         .63 &         .96 &             .62 &         .53 &         .59 &             .54 &         .40 &            2 &               13 &            6 &            1 &               97 &           55 &            9 &               83 &           42 \\
    \textsc{lime}$_{erase-1000}$ & .93 &             .63 &         .65 &         1. &             .77 &         .71 &         .88 &             .59 &         .55 &            1 &               13 &            4 &            1 &               95 &           32 &            3 &               78 &           31 \\
    \textsc{lime}$_{erase-3000}$ & .94 &             .63 &         .65 &         1. &             .78 &         .74 &         .94 &             .59 &         .58 &            1 &               13 &            4 &            1 &               96 &           27 &            1 &               78 &           28 \\
    \bottomrule
    \end{tabular}
    }
    \caption{Precision and rank of all the method configurations across the datasets and shortcut types for BERT.}
    \label{tab:results:bert}
\end{table*}

\end{document}